\pdfoutput=1
\documentclass[11pt]{article}

\usepackage[final]{acl}

\usepackage{times}
\usepackage{latexsym}
\usepackage{amsmath}
\usepackage{booktabs}
\usepackage{makecell}

\usepackage{xspace}
\usepackage{bm}
\usepackage{xcolor}

\usepackage[T1]{fontenc}
\usepackage[utf8]{inputenc}

\usepackage{microtype}

\usepackage{inconsolata}

\usepackage{graphicx}

\title{\textsc{GT2Vec}: Large Language Models as Multi-Modal Encoders for \\ Text and Graph-Structured Data}

\author{
  \textbf{Jiacheng Lin\textsuperscript{1}},
  \textbf{Kun Qian\textsuperscript{2}},
  \textbf{Haoyu Han\textsuperscript{3}},
  \textbf{Nurendra Choudhary\textsuperscript{2}},
\\
  \textbf{Tianxin Wei\textsuperscript{1}},
  \textbf{Zhongruo Wang\textsuperscript{2}},
  \textbf{Sahika Genc\textsuperscript{2}},
  \textbf{Edward W. Huang\textsuperscript{2}},
\\
  \textbf{Sheng Wang\textsuperscript{4}},
  \textbf{Karthik Subbian\textsuperscript{2}},
  \textbf{Danai Koutra\textsuperscript{5}},
  \textbf{Jimeng Sun\textsuperscript{1}}
\\
\\
  \textsuperscript{1}University of Illinois Urbana-Champaign,
  \textsuperscript{2}Amazon,
  \textsuperscript{3}Michigan State University,\\
  \textsuperscript{4}University of Washington,
  \textsuperscript{5}University of Michigan
\\
}

\def\ourmethod{\textsc{GT2Vec}\xspace}

\begin{document}
\maketitle
\begin{abstract}
Graph-structured information offers rich contextual information that can enhance language models by providing structured relationships and hierarchies, leading to more expressive embeddings for various applications such as retrieval, question answering, and classification. However, existing methods for integrating graph and text embeddings, often based on Multi-layer Perceptrons (MLPs) or shallow transformers, are limited in their ability to fully exploit the heterogeneous nature of these modalities. To overcome this, we propose \ourmethod, a simple yet effective framework that leverages Large Language Models (LLMs) to jointly encode text and graph data. Specifically, \ourmethod employs an MLP adapter to project graph embeddings into the same space as text embeddings, allowing the LLM to process both modalities jointly. Unlike prior work, we also introduce contrastive learning to align the graph and text spaces more effectively, thereby improving the quality of learned joint embeddings. Empirical results across six datasets spanning three tasks—knowledge graph-contextualized question answering, graph-text pair classification, and retrieval—demonstrate that \ourmethod consistently outperforms existing baselines, achieving significant improvements across multiple datasets. These results highlight \ourmethod's effectiveness in integrating graph and text data. Ablation studies further validate the effectiveness of our method.
\end{abstract}

\section{Introduction}
In the realm of natural language processing (NLP), text embeddings play a pivotal role by transforming textual information into numerical representations, which facilitate a multitude of machine learning applications on the Web, such as question answering (QA) \cite{DBLP:conf/naacl/DevlinCLT19, DBLP:conf/emnlp/ReimersG19}, retrieval \cite{DBLP:conf/emnlp/Ni0LDAMZLHCY22, wang2022text, DBLP:conf/sigir/MaWYWL24, choi2021evaluation}, and classification tasks \cite{wang2022text, DBLP:conf/naacl/DevlinCLT19}. These applications can benefit from the integration of graph-structured data to enhance the capabilities of NLP systems, by providing contextual information or by augmenting the original tasks with additional information. For example, prior work has shown that a QA system that includes a knowledge graph as input can leverage the relationships and hierarchies within the graph to more accurately understand and handle complex queries \cite{DBLP:conf/naacl/YasunagaRBLL21, zhang2022greaselm}. To effectively integrate these two modalities, it is essential to develop methods to learn joint embeddings of graph-structured data and text data. Such embeddings can provide a unified representation that captures important information from both modalities, leading to performance improvements across various NLP tasks.

Prior research has introduced various ways to learn joint embeddings of text and graph-structured data for embedding tasks~%in the NLP field 
\cite{DBLP:conf/naacl/YasunagaRBLL21, DBLP:conf/emnlp/FengCLWYR20, zhang2022greaselm, DBLP:conf/emnlp/LinCCR19}. These methods typically utilize either a Multi-layer Perceptron (MLP) or a shallow transformer \cite{DBLP:conf/nips/VaswaniSPUJGKP17} to integrate text features and graph embeddings encoded respectively by language models (LMs) \cite{DBLP:conf/naacl/DevlinCLT19, DBLP:journals/corr/abs-1907-11692} and graph neural networks (GNNs) \cite{DBLP:conf/iclr/KipfW17, DBLP:conf/iclr/VelickovicCCRLB18, DBLP:conf/iclr/XuHLJ19}. Despite their effectiveness, these approaches demonstrate a restricted capacity to fuse the features of the two modalities. The primary limitation arises from the limited ability of MLPs and shallow transformers to manage the high-dimensional and heterogeneous nature of joint embeddings, which can result in sub-optimal utilization of the rich contextual information from text and graph data. 
%Consequently, there is a growing need to explore more advanced integration techniques that can more effectively harness the complementary strengths of text and graph-structured data, thereby enhancing the overall robustness and accuracy of NLP systems.
%
Recently, large language models (LLMs) have demonstrated significant potential for integrating and understanding modalities beyond just text. A representative example of this capability is in Vision-Language Models (VLMs) \cite{DBLP:conf/nips/LiuLWL23a, DBLP:conf/iclr/Zhu0SLE24, DBLP:journals/corr/abs-2310-09478, DBLP:conf/icml/0008LSH23}, where visual tokens are combined with textual input and processed together by LLMs. This integration leverages the powerful capabilities of LLMs to handle multimodal data, allowing for a more holistic understanding of content that spans different forms of information. Inspired by this, we explore the potential of employing LLMs to better integrate text and graph-structured data, aiming to overcome the limitations observed from current approaches.

In this paper, we present \ourmethod, a simple yet effective framework to learn joint embeddings of text and graph data, leveraging the advanced capabilities of LLMs to address the limitations of previous approaches. Our method seamlessly integrates graph and text embeddings within the LLM framework, enhancing their alignment and interaction. Specifically, we transform graph embeddings into the same space as text embeddings using a multi-layer perceptron (MLP) adapter, enabling the LLM to process both modalities together. Additionally, we propose a contrastive learning strategy to better align the graph and text spaces, ensuring that the model learns richer representations of the combined data. Our extensive empirical analysis across a variety of NLP tasks highlights the key advantage of \ourmethod: the ability to leverage LLMs' strong language understanding and reasoning capabilities to process multimodal data, thus providing a more holistic and nuanced representation of both graph-structured and textual information. This integration leads to significant improvements in various NLP tasks, demonstrating the potential of \ourmethod to push the boundaries of joint text and graph-embedding techniques. 

Our contributions are summarized as follows:
\begin{itemize}
\item \textbf{Integration of LLMs for Joint Embeddings}: We propose \ourmethod framework that leverages the strengths of LLMs to align and integrate graph and text embeddings. \ourmethod effectively captures the rich contextual information of both modalities, enabling more robust joint representations.
\item \textbf{Contrastive Learning for Graph-Text Alignment}: We introduce a contrastive learning mechanism to explicitly align graph and text embeddings, enabling the model to better integrate the two modalities. 

% This alignment enhances the quality of joint representations, leading to improved performance in downstream tasks.
\item \textbf{Extensive Empirical Validation}: We conduct extensive experiments on six datasets spanning three different tasks: knowledge graph (KG)-contextualized QA, graph-text pair classification, and retrieval tasks. \ourmethod achieves superior performance on all the three tasks, demonstrating its ability to effectively integrate graph and text data for enhanced multi-modal representation learning.
\end{itemize}

% Especially on the KG-contextualized QA tasks, we found that our \ourmethod achieved an absolute accuracy improvement of 6.89\% on CommonsenseQA, 9.46\% on OpenBookQA, and 11.4\% on MedQA. In graph-text pair classification and retrieval tasks, 

%To evaluate \ourmethod, we conduct extensive experiments on six datasets spanning three different tasks: knowledge graph (KG)-contextualized QA, graph-text pair classification, and retrieval tasks. Especially on KG-contextualized QA, we found that our \ourmethod achieved an absolute accuracy improvement of 6.57\% on CommonsenseQA, 9.46\% on OpenBookQA, and 11.4\% on MedQA. In graph-text pair classification and retrieval tasks, \ourmethod also achieves superior performance by effectively integrating and aligning graph and text embeddings. The results of our experiments highlight the key advantage of \ourmethod: the ability to leverage LLMs' strong language understanding and reasoning capabilities to process multimodal data, thus providing a more holistic and nuanced representation of both graph-structured and textual information. This integration leads to significant improvements in various NLP tasks, demonstrating the potential of \ourmethod to push the boundaries of joint text and graph-embedding techniques.

\begin{figure*}
    \centering
    \includegraphics[width=0.85\linewidth]{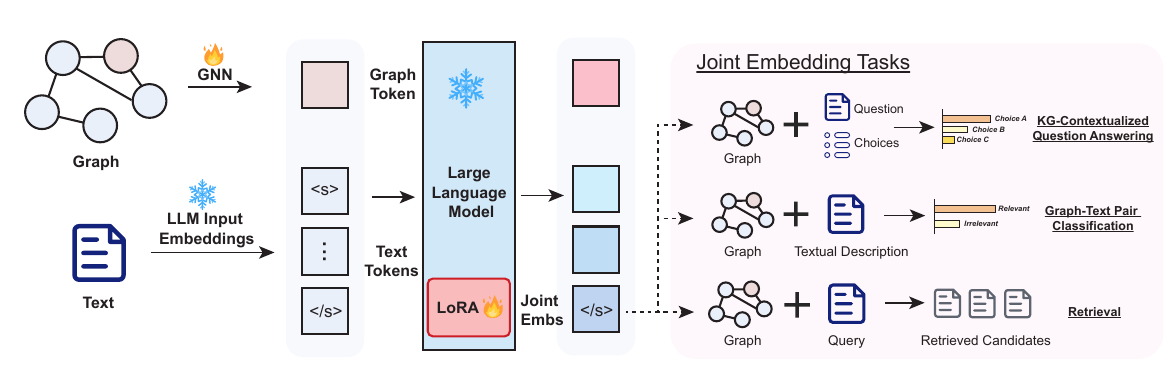}
    \vspace{-0.3cm}
    \caption{Overview of \ourmethod framework. Unlike the common use of LLMs for generation tasks, we leverage LLMs to obtain joint embeddings of both text and graph data. We encode the input graph with a GNN, which provides the graph embeddings. The graph embeddings are then transformed into the word embedding space in the large language model. These embeddings are then fed into a large language model, and the outputs are utilized for various downstream tasks. }
    \vspace{-0.3cm}
    \label{fig:framework}
\end{figure*}

\section{Problem Statement}
Given an input text $x$ and its corresponding graph context $\mathcal{G}$, where $\mathcal{G} = \{\mathcal{V}, \mathcal{E}\}$ consists of a set of nodes $\mathcal{V}$ and edges $\mathcal{E} \subseteq \mathcal{V} \times \mathcal{R} \times \mathcal{V}$ that connect nodes via relationships $\mathcal{R}$, our goal is to extract joint embeddings $\phi(x, \mathcal{G})$. Here, $\phi$ is a learned function that maps the input text $x$ and graph $\mathcal{G}$ into a unified vector representation, capturing the multimodal information. These joint embeddings $\phi(x, \mathcal{G})$ are then used for downstream tasks. In this paper, we focus on three specific downstream tasks: multi-choice QA contextualized by KG, graph-text pair classification, and retrieval tasks.

% Given an input text $x$ and graph context $\mathcal{G}$ related to $x$, our goal is to extract joint embeddings $\phi(x, \mathcal{G})$ that can be used for downstream multi-modal embedding tasks. Here, $\phi$ is a learned function to map the input text $x$ and graph $\mathcal{G}$ into a vector. In this paper, we focus on three downstream tasks, i.e., multi-choice QA contextualized by knowledge graph (KG), graph-text pair classification, and retrieval tasks. 
% Given a graph input $\mathcal{G} = \{\mathcal{V}, \mathcal{E}\}$ where $\mathcal{V}$ represents the set of nodes and $\mathcal{E} \subseteq \mathcal{V} \times \mathcal{R} \times \mathcal{V}$ denotes the edges that connect nodes within in $\mathcal{V}$ via relationships defined in $\mathcal{R}$,

\subsection{KG-Contextualized QA}
Given a question $q$ in text form, KG context $\mathcal{G}$, and answer candidate set with $n$ choices $\bm{a} = \{a_1, \cdots, a_n\}$, KG-contextualized QA tasks aim to find the correct textual answer $a_i$ from $\bm{a}$. Each choice $a_i$ is first concatenated with the question $q$, leading to the input $x = [q, a_i]$, where $[\cdot, \cdot]$ denotes the concatenation of text. We then extract the joint embeddings $\phi(x, \mathcal{G})$ which is fed into a MLP layer to calculate scores. The choice with the highest score is selected as the prediction.

\subsection{Graph-Text Pair Classification}
In the task of graph-text pair classification, the objective is to determine the relevance between a given graph, represented as $\mathcal{G}$, and a corresponding textual description $x$. This involves assessing whether the content and structure of $\mathcal{G}$ are accurately reflected or described by $x$.

To achieve this, we first compute the joint embeddings $\phi(x, \mathcal{G})$ which capture the features and relationships contained in both the text and the graph. Once the joint embeddings are obtained, they are input into a MLP classifier. It outputs a prediction score which measures the likelihood of the graph $\mathcal{G}$ matching the text $x$. 

The significance of this task lies in its ability to improve the quality of training data for tasks such as generating text from knowledge graphs and vice versa \cite{DBLP:conf/acl/KeJRCWSZH21, DBLP:journals/semweb/LehmannIJJKMHMK15, DBLP:conf/coling/ColasAW22}. By accurately classifying graph-text pairs, the model can help reduce noise in training data, which in turn improves the overall performance of generation tasks \cite{DBLP:journals/semweb/LehmannIJJKMHMK15}.

% Given an input graph $\mathcal{G}$ and a text description $x$, we aim to verify whether the input graph $\mathcal{G}$ and text $x$ match. Specifically, after obtaining the joint embeddings $\phi(x, \mathcal{{G}})$, we leverage a MLP classifier to output the prediction. 

% In this paper, we consider the molecule property prediction task for evaluation. Specifically, the input graph $\mathcal{G}$ denotes the molecule and the text $x$ describes a certain molecule property. The goal is to verify whether molecule $\mathcal{G}$ has the property described in the text $x$.

\subsection{Retrieval}
% Given an input text $x$ and graph context $\mathcal{G}$, our goal is to extract 
For the retrieval task, given a textual query $q$ accompanied by its graph context $\mathcal{G}_q$, the goal is to retrieve the most relevant candidate from a set of text-based options. Each candidate $c_i$ in the candidate set $\bm{c} = \{c_1, \cdots, c_m\}$ also has an associated graph context $\mathcal{G}_{c_i}$. The task involves comparing the query-graph pair $(q, \mathcal{G}_q)$ against each candidate-graph pair $(c_i, \mathcal{G}_{c_i})$.

To achieve this, we first generate the joint embeddings $\phi(q, \mathcal{G}_q)$ for the query and $\phi(c_i, \mathcal{G}_{c_i})$ for each candidate. These embeddings encapsulate the features and relationships pertinent to their respective texts and graphs. The cosine similarity between the embeddings of the query and each candidate is calculated and is then used for the selection of the most relevant options. 
\section{\ourmethod}
In this work, we propose a simple yet effective framework \ourmethod to learn joint embeddings of text and graphs, as illustrated in Figure \ref{fig:framework}. Specifically, graph-structured data are first extracted into a graph token, which is then fed into the LLM backbone together with the text tokens (\S\ref{subsec:graph_data_encoding}, \S\ref{subsec:fusion_graph_text}). The LLM backbone outputs the joint embeddings, which can be used for the downstream tasks, such as classification and retrieval. 

A key aspect of \ourmethod is the explicit alignment between the graph and text embeddings. This alignment is crucial, as it allows the LLM backbone to better integrate the structured knowledge from the graph with the unstructured text, improving the quality of the joint representations (\S\ref{sec:exp}). To achieve this, we introduce a contrastive learning mechanism that explicitly maps embeddings from both modalities into a shared space (\S \ref{subsec:alignment}).

While LLMs are commonly used for generation tasks in an auto-regressive paradigm, \ourmethod takes a different route by utilizing LLMs as powerful encoders. This allows us to directly obtain robust joint embeddings of text and graph data by leveraging the rich contextual understanding of LLMs.

Our proposed architecture, \ourmethod, consists of three main components: a graph encoder, an MLP adapter, and a large language model backbone. Below, we discuss the components’ design and implementation details, along with the alignment mechanism.

\subsection{Graph Data Encoding}
\label{subsec:graph_data_encoding}
\ourmethod employs a graph encoder to obtain graph embeddings, which encapsulate essential information extracted from the contextual structure of the graph. Next, we describe the two-step process, which includes: (1) the integration of the query node in the graph; and (2) the graph encoding. 

\subsubsection{Query Node Integration into Graph Structures} 
We first initialize node embeddings within the input graph $\mathcal{G}$ with a language model (e.g., RoBERTa \cite{DBLP:journals/corr/abs-1907-11692}). Following prior work \cite{lin2019kagnet, DBLP:conf/naacl/YasunagaRBLL21, zhang2022greaselm}, we link entities mentioned in the query to nodes in the graph, denoting these nodes as $\mathcal{V}_{{\rm linked}}$. We then introduce a new node termed the query node, denoted as $v_q$. This query node is initialized using a language model by encoding the input query text. The query node $v_q$ is then connected to all nodes within $\mathcal{V}_{{\rm linked}}$, enhancing the connection between the query and the nodes within the graph. We denote the updated graph with $\mathcal{G}' = \{\mathcal{V}', \mathcal{E}'\}$.

\subsubsection{Graph Encoding Process}
\label{subsubect:graph_encode}
The updated graph $\mathcal{G}'$ is then fed into an encoder for feature extraction. We adopt a modified version of graph attention network (GAT) \cite{DBLP:conf/iclr/VelickovicCCRLB18, DBLP:conf/naacl/YasunagaRBLL21, zhang2022greaselm} as the graph encoder. Specifically, in each layer of GAT, the message-passing process is formulated as
\begin{equation}
    \resizebox{0.85\hsize}{!}{$
    \bm{h}_v^{(\ell+1)} = f_n \left( \sum_{s \in \mathcal{N}_v \cup \{v\}} \alpha_{sv} \mathbf{m}_{sv} \right) + \bm{h}_v^{(\ell)}
    $}
\end{equation}
where $\mathcal{N}_v$ denotes the neighbors of node $v$, $\mathbf{m}_{sv}$ represents the message from each neighbor node $s$ to node $v$. $\alpha_{sv}$ is the attention weight. $f_n$ is a 2-layer MLP. The messages $\mathbf{m}_{sv}$ from node $s$ to $v$ are computed as the following:
\begin{align}
    \resizebox{0.85\hsize}{!}{$
    \bm{r}_{sv} = f_r(\bm{e}_{sv}, \bm{u}_s, \bm{u}_v)~~~~~ \mathbf{m}_{sv} = f_m(\bm{h}_v^{(\ell)}, \bm{u}_v, \bm{r}_{sv})$}
\end{align}
where $\bm{u}_s, \bm{u}_v$ denotes node type embeddings, and $\bm{e}_{sv}$ is edge embeddings, $f_r$ is a 2-layer MLP, and $f_m$ is a linear projection. Additionally, the attention weight $\alpha_{sv}$, which measures the importance of each neighbor's message, is calculated in the following manner:
\begin{align}
    \resizebox{0.8\hsize}{!}{$
    \bm{q}_s = f_q(\bm{h}_s^{(\ell)}, \bm{u}_s) ~~~~~~ \bm{k}_v = f_k(\bm{h}_v^{(\ell)}, \bm{u}_v, \bm{r}_{sv})$}\\
    \resizebox{0.8\hsize}{!}{$
    \gamma_{sv} = \cfrac{\bm{q}_s^{\top} \bm{k}_v}{\sqrt{D}} ~~~~~~ \alpha_{sv} = \cfrac{\exp(\gamma_{sv})}{\sum_{v' \in \mathcal{N}_{s} \cup \{s\}}\exp(\gamma_{sv'})}$}
\end{align}
where $f_q$ and $f_k$ are linear transformation functions, and $D$ is the hidden dimension. Following $L$ layers of message passing, we concatenate the final layer embeddings of query node $v_q$, the average pooling of the node embeddings in the final layer, and the text embeddings of the input query, then employ an MLP to generate the graph embeddings $\bm{g}$.

\subsection{Fusion of Graph and Textual Data in LLM}
\label{subsec:fusion_graph_text}
We then integrate the graph embeddings with textual information using the LLM to produce embeddings suitable for downstream tasks. To facilitate this integration, we draw inspiration from practices in computer vision \cite{DBLP:conf/nips/LiuLWL23a, DBLP:conf/iclr/Zhu0SLE24}, where image embeddings are first transformed into the text space to be processed by LLMs. Similarly, we also convert the graph embeddings into the text space. We employ an MLP adapter for this purpose, which projects the graph token embeddings into the language model space using a two-layer MLP with ReLU activation functions, leading to the transformed graph embeddings $\tilde{\bm{g}}$. The transformation allows us to insert the processed graph token at the beginning of the text sequence, formatted as $[\text{\textit{graph token}}, \text{<$s$>}, token\text{ 1}, token\text{ 2}, ..., \text{<$/s$>}]$. In this sequence, the graph token, output by the MLP adapter, precedes the textual tokens, which are derived from the initial input embeddings of the LLM. This arrangement ensures that the initial context for the LLM processing includes both graph-derived and textual information. 

We then feed this sequence into the LLM. The embeddings of the $\text{<$/s$>}$ token from the last layer of the LLM are considered as the final output embeddings $\bm{z}$, encapsulating the combined knowledge of the graph and text inputs. This integration process leverages the LLM's capacity to synthesize information across different modalities, optimizing the embeddings for subsequent applications.

% \textbf{MLP adapter.} We aim to project graph token embeddings into the language model space. To this end, we use two-layer MLP with ReLU activation functions. We then insert the transformed graph token before tokens of input text, i.e.:

% \textbf{Large language model.} Our \ourmethod framework is compatible with any off-the-shelf LLMs. Here, we conducted experiments with Mistral-7B-Instruct (decoder) \cite{jiang2023mistral} and E5-Mistral (encoder). Here, E5-Mistral is a powerful text encoder customized for text embedding tasks fine-tuned from Mistral-7B-Base by contrastive learning. Moreover, we applied bidirectional attention mask instead of casual mask when using the decoder version model Mistral-7B-Instruct. In this case, the bidirectional attention mask allows for more comprehensive integration of context from both front and behind tokens, enhancing the model's ability to obtain the embeddings based on the full scope of the input text. We view the embeddings of </s> token in the last layer as the output embeddings.

\subsection{Graph-Text Alignment via Contrastive Learning}
\label{subsec:alignment}
\begin{figure}
    \centering
    \includegraphics[width=0.5\textwidth]{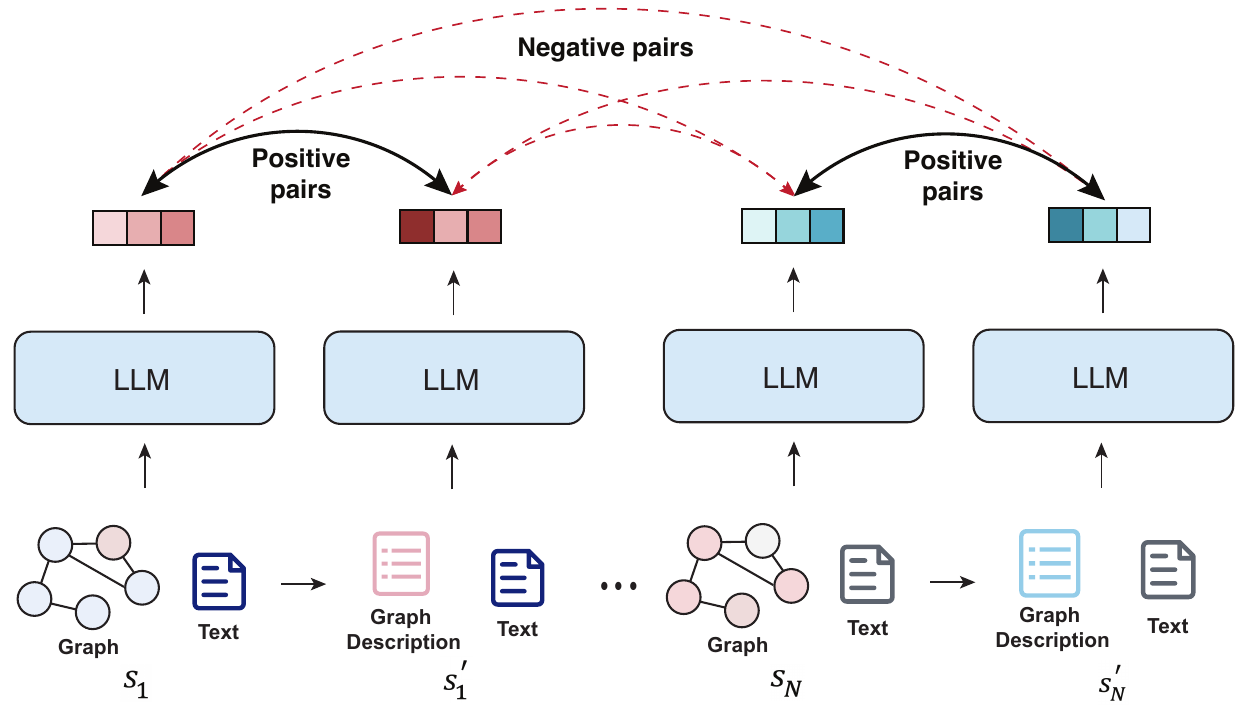}
    \caption{Overview of graph-text alignment through contrastive learning.}
    \vspace{-1em}
    \label{fig:alignment}
\end{figure}

 % Originally, each data input consists of a graph and its corresponding text. To align the graph space with the text space, each graph is converted into a language-based text description, leading to two processing streams: the original graph + text, and the new graph description + text. We then leverage contrastive learning where transformed versions of the same sample serve as positive pairs, while different samples within the batch act as negative pairs.

As we mentioned above, in \ourmethod, graph embeddings are converted into the text space via a MLP module and then are processed by the LLM together with the input text tokens. To better align the graph embedding space and text space, we consider contrastive learning for the graph-text alignment. 

Contrastive learning has been broadly used in various domains \cite{he2020momentum, chen2020simple, linpisces, DBLP:conf/iclr/LinXXW24, yu2022graph, DBLP:conf/acl/IterGLJ20, xu2022pisces}. The key idea behind contrastive learning is to minimize the distance between similar or positive pairs in the embedding space, while maximizing the separation between dissimilar or negative pairs. Although similar techniques have been applied in graph representation learning \cite{yu2022graph, you2020graph, fang2024gaugllm} and language models \cite{iter2020pretraining, fang2020cert, giorgi2021declutr}, these approaches typically operate within a single modality (either graphs or text). In contrast, our \ourmethod framework aligns the embedding spaces between text and graphs by (1) converting graph data into textual descriptions and (2) using contrastive learning to strengthen the connections between the textual descriptions of graphs and their corresponding original graph representations, as shown in Figure \ref{fig:alignment}.

\subsubsection{Translating Graphs into Textual Descriptions}
% In Section \ref{subsubect:graph_encode}, we have mentioned that graph embeddings in \ourmethod are obtained through average pooling, which involves summarizing the nodes to form a comprehensive representation of the entire graph \cite{chen2022distribution, pang2021graph}. 

We first align the graph embeddings with text embeddings derived from descriptions of the same graph. Specifically, we start by listing all triples within the graph in a simple format: [entity (node), relation (edge), entity (node)]. This textual representation of the graph is then converted into a textual format by reorganizing the entities and relations into natural language. For example, the triples "\textit{(analyzing, causes, new knowledge), (knowledge, causes, learn), (learn, causes, find information)}" are transformed into the following textual description: "\textit{analyzing causes new knowledge; knowledge causes learn; learn new causes find information.}" 

% The generated text descriptions are then truncated to a length specified by $L$. 

% The rationale for this selective filtering mirrors the processing during the graph embedding phase, where a graph attention network is utilized. The attention mechanism in the graph encoder selectively weighs certain nodes, inspiring us that employing a selective process for the generated text descriptions could enhance alignment. This alignment is crucial for applying contrastive learning effectively.

% [Graph embeddings are obtained through a pooling process, which involves summarizing the nodes to form a comprehensive representation of the entire graph \cite{chen2022distribution, pang2021graph}. Based on this, we consider using summaries of texts to align with the graph embeddings. Specifically, we not only retain the original process that handles the graph embeddings but also introduce a new view: converting the graph into a natural language description. This description acts as a natural language summary of the input graph and can also be considered a form of embedding.]

\subsubsection{Contrastive Learning for Alignment}
Now, we have two branches in the \ourmethod framework: the original graph + text branch and the graph description + text branch. These branches create a dual-view architecture that facilitates the alignment of graph and text embedding spaces through contrastive learning.

In the contrastive learning stage, we define positive pairs as the embeddings from the branch processing the original graph and its corresponding textual description that accurately reflect the same information. Negative pairs, on the other hand, consist of unrelated graph-text pairs from within the same batch, ensuring that the model learns to distinguish between semantically aligned and unaligned graph-text pair data. Specifically, for the original graph + text branch, we use the way mentioned in \S \ref{subsec:fusion_graph_text} to generate the joint embeddings $\bm{z}_{\rm orig}$. This involves combining graph embeddings, obtained from a graph encoder, with textual embeddings generated by the LLM’s embedding layer to produce joint embeddings. For the graph description + text branch, we treat the input as pure text and use only the LLM to process it and generate the joint embeddings $\bm{z}_{\rm new}$. By applying infoNCE loss \cite{oord2018representation}, we have
\begin{equation}
\resizebox{0.96\hsize}{!}{$
\mathcal{L}^{(\text {infoNCE})}=-\sum_{i=1}^{n_{\text {batch}}} \log \left(\frac{\exp \left(\tilde{\boldsymbol{z}}_{\rm orig}^{\mathrm{T}}(i) \cdot \tilde{\boldsymbol{z}}_{\rm new}(i)\right)}{\sum_{j=1}^{n_{\text {batch}}} \exp \left(\tilde{\boldsymbol{z}}_{\rm orig}^{\mathrm{T}}(i) \cdot \tilde{\boldsymbol{z}}_{\rm new}(j)\right)}\right)
$}
\end{equation}
where $n_{\text {batch }}$ represents the number of samples in a training batch, and $\tilde{\boldsymbol{z}}_{\text {orig}}$ and $\tilde{\boldsymbol{z}}_{\text {new}}$ denote the normalized vectors of $\boldsymbol{z}_{\text {orig}}$ and $\boldsymbol{z}_{\mathrm{new}}$, respectively. As we present in  \S\ref{sec:exp}, the contrastive learning method improves \ourmethod's ability to better align the graph and text embeddings, which is further beneficial to the downstream tasks.

\begin{table*}[t]
    \centering
    \caption{Test accuracy comparison on CommonsenseQA and OpenBookQA. The baseline results are mainly sourced from \citet{zhang2022greaselm} and \citet{DBLP:conf/naacl/YasunagaRBLL21}. \textbf{Bold} indicates the best result, and \underline{underline} indicates the second best. LP means linear probe.}
    \resizebox{0.85\hsize}{!}{
    \begin{tabular}{lccc}
        \toprule
        \textbf{Methods} & \textbf{CommonsenseQA} & \makecell{\textbf{OpenBookQA} \\ \textbf{(w/o Scientific Facts)}} & \makecell{\textbf{OpenBookQA} \\ \textbf{(w/ Scientific Facts)}} \\
        \midrule
        \multicolumn{3}{l}{\textit{\textbf{Language Models Only}}} \\
        RoBERTa-Large \cite{DBLP:journals/corr/abs-1907-11692} & 68.69 $\pm$ 0.56 & 64.80 $\pm$ 2.37 & 78.40 $\pm$ 1.64 \\
        E5-Mistral, LP  \cite{wang2023improving} & 69.49 $\pm$ 0.28 & 74.80 $\pm$ 0.35 & 81.67 $\pm$ 0.31 \\
        E5-Mistral, LoRA  \cite{wang2023improving} & 78.73 $\pm$ 0.16 & 85.60 $\pm$  0.20 & 91.87 $\pm$ 0.12 \\
        \midrule
        \multicolumn{3}{l}{\textit{\textbf{LM + KG}}} \\
        RGCN \cite{DBLP:conf/esws/SchlichtkrullKB18} & 68.41 $\pm$ 0.66 &  62.45 $\pm$ 1.57 & 74.60 $\pm$ 2.53 \\
        GconAttn \cite{DBLP:conf/aaai/WangKMYTACFMMW19} & 68.59 $\pm$ 0.39 & 64.75 $\pm$ 1.48 & 71.80 $\pm$ 1.21 \\
        % KagNet \cite{DBLP:conf/emnlp/LinCCR19} & 69.01 $\pm$ 0.22 & - & - \\
        % RN \cite{DBLP:conf/nips/SantoroRBMPBL17} & 69.08 $\pm$ 0.91 &  65.20 $\pm$ 1.18 & 75.35 $\pm$ 1.39 \\
        MHGRN \cite{DBLP:conf/emnlp/FengCLWYR20} & 71.11 $\pm$ 0.10 &  66.85 $\pm$ 1.19 & 81.87 $\pm$ 1.86 \\
        QA-GNN \cite{DBLP:conf/naacl/YasunagaRBLL21} & 73.41 $\pm$ 0.92 & 67.80 $\pm$ 2.75  & 82.77 $\pm$ 1.56 \\
        GreaseLM \cite{zhang2022greaselm} & 74.20 $\pm$ 0.40 & 65.60 $\pm$ 0.40 & 83.87 $\pm$ 1.29 \\
        \midrule
        {\ourmethod, LP (Ours)} & \underline{81.09} $\pm$ 0.73 & \underline{86.67} $\pm$ 1.10 & \underline{93.33} $\pm$ 0.42 \\
        {\ourmethod, LoRA (Ours)} & \textbf{81.39} $\pm$ 0.11 & \textbf{88.13} $\pm$ 0.42 & \textbf{93.67} $\pm$ 0.31 \\
        \bottomrule
    \end{tabular}
    }
    \vspace{-0.3em}
    \label{tab:csqa_obqa}
\end{table*}

\subsection{Training}
Our proposed framework, \ourmethod, can accommodate a wide array of tasks that require the integration of graph data and text. The training process for each task incorporates a task-specific loss function combined with a contrastive learning loss. The general form of the combined loss for each task can be represented as follows:
\begin{align}
    \label{eq:loss_comb}\mathcal{L} = \mathcal{L}^{(\rm task)} + \lambda \mathcal{L}^{(\text {infoNCE})}
\end{align}
where $\lambda$ is a hyperparameter used to adjust the weights between loss functions. $\mathcal{L}^{(\rm task)}$ refers to the loss directly associated with the primary objective of the task. For KG-Contextualized QA, we use cross-entropy loss; for graph-text pair classification, we apply binary cross-entropy (BCE) loss; and for retrieval, we use infoNCE loss. More details can be found in the Appendix \ref{appendix:training_objective}.

\section{Experiments}
\label{sec:exp}
Next, we assess the performance of \ourmethod and compare it to strong baselines on three types of tasks: KG-contextualized QA, graph-text pair classification, and retrieval tasks (\S \ref{subsec:qa}-\ref{subsec:retrieval}). We also perform extensive ablation studies to understand the impact of our design and other choices (\S \ref{subsec:ablation}).

\subsection{KG-contextualized QA Performance}
\label{subsec:qa}
\subsubsection{Datasets and Metrics}
% add some sentences for same datasets/splits

We first evaluated \ourmethod on three QA datasets, i.e., CommonsenseQA \cite{DBLP:conf/naacl/TalmorHLB19}, OpenBookQA \cite{DBLP:conf/emnlp/MihaylovCKS18}, and MedQA-USMLE \cite{DBLP:journals/corr/abs-2009-13081}. We split these datasets according to \citet{DBLP:conf/naacl/YasunagaRBLL21} and \citet{zhang2022greaselm} into train, validation, and test splits. For evaluation, we use accuracy as the metric to measure the performance on each dataset. For each question in the QA datasets, a subgraph context extracted from a KG is utilized to provide additional contextual information, following Yasunaga et al. \cite{DBLP:conf/naacl/YasunagaRBLL21} (see Appendix \ref{subsec:app_kqqa}). 

% To validate the effectiveness of our \ourmethod framework,
\subsubsection{Baselines} We compare \ourmethod with a range of baseline models, including both language models (LM) and hybrid approaches that integrate language models with knowledge graphs (LM+KG). 

% and AristoRoBERTa \cite{clark2020f}

\textbf{Fine-tuned LMs.} We consider the following vanilla fine-tuned language models (LMs): RoBERTa-Large \cite{DBLP:journals/corr/abs-1907-11692} and E5-Mistral \cite{wang2023improving}. Additionally, for MedQA-USMLE dataset, we use several domain-specific models, including SapBERT \cite{DBLP:conf/naacl/LiuSMBC21} and BioBERT \cite{lee2020biobert}. 

\textbf{Existing LM+KG models.} Our LM+KG baselines include: GreaseLM \cite{zhang2022greaselm} QAGNN \cite{DBLP:conf/naacl/YasunagaRBLL21}, Relation Network (RN) \cite{DBLP:conf/nips/SantoroRBMPBL17}, RGCN \cite{DBLP:conf/esws/SchlichtkrullKB18}, GconAttn \cite{DBLP:conf/aaai/WangKMYTACFMMW19}, and MHGRN \cite{DBLP:conf/emnlp/FengCLWYR20}. Unlike these models that leverage GNNs as the backbone, our framework \ourmethod adopts more powerful LLMs for better performance. We adopt E5-Mistral \cite{wang2023improving} as \ourmethod's LLM backbone.

\subsubsection{Training Details}
During training, we adopt parameter efficient finetuning techniques, including Linear Probe (LP) \cite{radford2021learning} and Low-Rank Adaptation (LoRA) \cite{DBLP:conf/iclr/HuSWALWWC22}, for E5-Mistral and \ourmethod. LP freezes the pre-trained LLM and trains a linear classifier on top, while LoRA introduces low-rank adaptation layers for efficient finetuning. we employ full parameter fine-tuning for the other baselines. Further details on the training process such as hyperparameters can be found in Appendix \ref{app:section:exp}.

% During the training process, we freeze the LLM part in \ourmethod, only training GNNs and the MLP adapter. In this way, we reduced the computational resources required while still leveraging LLMs' robust language understanding capabilities. 

\begin{table}[t]
    \centering
    \caption{Test accuracy comparison on MedQA-USMLE. The baseline results are mainly sourced from \citet{zhang2022greaselm}. \textbf{Bold} indicates the best result, and \underline{underline} indicates the second best. LP means linear probe.}
    \resizebox{0.9\hsize}{!}{
    \begin{tabular}{lc}
        \toprule
        \textbf{Methods} & \textbf{Test Acc} \\
        \midrule
        \multicolumn{1}{l}{\textit{\textbf{Language Models Only}}} \hspace{4mm} \\
        BERT-Base  \cite{DBLP:conf/naacl/DevlinCLT19} & 34.3 \\
        BioBERT-Base  \cite{lee2020biobert} & 34.1 \\
        RoBERTa-Large  \cite{DBLP:journals/corr/abs-1907-11692} & 35 \\
        BioBERT-Large  \cite{lee2020biobert} & 36.7 \\
        SapBERT  \cite{DBLP:conf/naacl/LiuSMBC21} & 37.2 \\
        E5-Mistral, LP  \cite{wang2023improving} & 39.4 $\pm$ 1.1 \\
        E5-Mistral, LoRA \cite{wang2023improving} & \underline{51.1} $\pm$ 0.3 \\
        \midrule
        \multicolumn{2}{l}{\textit{\textbf{LM + KG}}} \\
        QA-GNN \cite{DBLP:conf/naacl/YasunagaRBLL21} & 38 \\
        GreaseLM \cite{zhang2022greaselm} & 38.5 \\
        \midrule
        {\ourmethod, LP (Ours)} & {49.9} $\pm$ 0.9 \\
        {\ourmethod, LoRA (Ours)} & \textbf{53.4} $\pm$ 0.3  \\
        \bottomrule
    \end{tabular}
    }
    \label{tab:medqa}
    \vspace{-1em}
\end{table}

\subsubsection{Results}

% All baseline results are sourced from \cite{DBLP:conf/naacl/YasunagaRBLL21, zhang2022greaselm}, except for E5-Mistral, whose results were obtained from our experiments.

We first conduct comparison experiments on CommonsenseQA and OpenBookQA datasets, as illustrated in Table \ref{tab:csqa_obqa}. In both datasets, our framework outperforms all other methods significantly. Specifically, on CommonsenseQA, it surpasses the strongest baseline E5-Mistral, achieving a test accuracy of 81.39\%.  On OpenBookQA, \ourmethod also demonstrates superior performance, achieving 88.13\% test accuracy. When further integrating the extra corpus of scientific facts provided by OpenbookQA \cite{clark2020f}, our model reaches a remarkable 93.67\% accuracy, outperforming all the baseline models that also utilized scientific facts. The remarkable performance can be attributed primarily to the robust capabilities of the LLM backbone integrated within our framework. The LLM backbone not only enhances language understanding but also integrates the contextual knowledge derived from graphs, thereby substantially improving KG-contextualized QA performance.

% These results clearly demonstrate the exceptional capability of \ourmethod, with the LLM backbone playing a critical role in significantly enhancing both language understanding and the integration of external knowledge.

% 

% The significant boost in performance can be attributed to the LLM backbone's ability to handle complex medical terminology and integrate knowledge from medical knowledge graphs effectively.

We further evaluate \ourmethod on the MedQA-USMLE dataset to assess its generalization capability across specialized domains and report the results in Table \ref{tab:medqa}. \ourmethod achieves a test accuracy of 53.4\%, outperforming all the baselines including domain-specific models. This result further reinforces the adaptability of \ourmethod across different domains showcasing its ability to excel not only in general and scientific question answering but also in knowledge-intensive and highly specialized fields.

% In addition to evaluating \ourmethod against various baseline models, we also conduct a direct comparison between \ourmethod and its LLM backbone, E5-Mistral \cite{wang2023improving}. For a fair comparison, we keep the E5-Mistral frozen, training only the classification head, which is consistent with how we evaluate \ourmethod. This ensures that any performance differences can be attributed to the integration of graph knowledge rather than differences in backbone training. The experimental results demonstrate that \ourmethod significantly outperforms the standalone E5-Mistral model across all datasets by 11-12\%. These results clearly highlight the benefits of integrating structured knowledge graphs with the LLM backbone. While the LLM backbone alone captures rich linguistic information, the additional graph context enables the model to better understand the relationships and structured information relevant to the task, resulting in significant performance improvements across a wide range of QA tasks.

% This task requires the model to determine whether a given text accurately corresponds to graph-structured data.
\subsection{Graph-Text Pair Classification Performance}
Building upon the strong results achieved in the KG-contextualized QA tasks, we further evaluate \ourmethod in a different task: graph-text pair classification. Our objective in evaluating this task is to assess \ourmethod's ability to generalize beyond QA scenarios, demonstrating its versatility in handling a broader range of graph-text embedding tasks. We evaluate models on the WebNLG \cite{web_nlg} dataset and use accuracy as the evaluation metric, measuring the proportion of correctly classified graph-text pairs.

% \subsubsection{Datasets and Metrics}
% We use a dataset derived from WebNLG \cite{web_nlg}, which contains 10,000 training, 4,000 validation, and 2,000 test graph-text pairs. Each graph consists of triples (entity, relation, entity) from DBPedia \cite{DBLP:journals/semweb/LehmannIJJKMHMK15}, and each pair is labeled as 1 if the text correctly describes the entities and their relationships, and 0 otherwise. Each of the train, validation, and test splits has an equal number of positive and negative samples. We use accuracy as the evaluation metric, measuring the proportion of correctly classified graph-text pairs. 

% The details of how we curated this dataset can be found in the Appendix \ref{appendix:subsec_webnlg}. 

% Unless otherwise specified, we refer to this dataset simply as WebNLG throughout the paper.

% \subsubsection{Baselines}
% We compare \ourmethod against strong KG+LM baselines, including RGCN \cite{DBLP:conf/esws/SchlichtkrullKB18}, MHGRN \cite{DBLP:conf/emnlp/FengCLWYR20}, GreaseLM \cite{zhang2022greaselm}, and QA-GNN \cite{DBLP:conf/naacl/YasunagaRBLL21}. These models have demonstrated strong performance in QA tasks and provide open-source implementations, allowing us to reliably compare their performance against \ourmethod.

\begin{table}[t]
    \centering
    \caption{Test accuracy comparison on WebNLG dataset.}
    \resizebox{0.8\hsize}{!}{
    \begin{tabular}{lc}
        \toprule
        \textbf{Methods} & \textbf{Test Acc} \\
        \midrule
        \multicolumn{1}{l}{\textit{\textbf{LM + KG}}} \hspace{7mm} \\
        RGCN \cite{DBLP:conf/esws/SchlichtkrullKB18} & 63.20 $\pm$ 0.49 \\
        MHGRN   \cite{DBLP:conf/emnlp/FengCLWYR20} &  84.98 $\pm$ 0.53 \\
        QA-GNN \cite{DBLP:conf/naacl/YasunagaRBLL21} & 75.55 $\pm$ 3.54 \\
        GreaseLM \cite{zhang2022greaselm} & 82.50 $\pm$ 4.29  \\
        \midrule
        % Janus (Mistral-7B-Instruct) & 83.6 \\
        {\ourmethod, LP (Ours)} & \underline{88.43} $\pm$ 1.33 \\
        {\ourmethod, LoRA (Ours)} & \textbf{89.70} $\pm$ 0.34 \\
        \bottomrule
    \end{tabular}
    }
    % \vspace{-1em}
    \label{tab:webnlg}
\end{table}

% more stale than others (GreaseLM)

% comments for high variance

The results of the graph-text pair classification task are presented in Table \ref{tab:webnlg}. \ourmethod significantly outperforms all baseline models by at least 4.72\%, achieving a test accuracy of 89.70\%. While models like GreaseLM and MHGRN perform well by incorporating knowledge graphs, they lack the deeper contextual understanding due to the limitations of their less powerful backbone models, such as shallow transformers or GNNs.

\subsection{Retrieval Performance}
\label{subsec:retrieval}
Next, we further evaluate \ourmethod's performance on retrieval tasks, where the objective is to retrieve relevant sentences or documents based on a query with graph-context. We conduct experiments on two datasets, SciFact \cite{DBLP:conf/emnlp/WaddenLLWZCH20} and FiQA \cite{DBLP:conf/www/MaiaHFDMZB18}, and report NDCG@10 as the primary evaluation metric. 
% \subsubsection{Datasets and Metrics}

% \subsubsection{Baselines}
% We compare \ourmethod against a variety of baselines, including both language models alone (e.g., RoBERTa \cite{DBLP:journals/corr/abs-1907-11692}, E5 \cite{wang2022text,wang2023improving}, GTR-XXL \cite{ni2022large}) and models that integrate knowledge graphs (e.g., QA-GNN \cite{DBLP:conf/naacl/YasunagaRBLL21}, GreaseLM \cite{zhang2022greaselm}). 

% \subsubsection{Results}

\begin{table}
    \centering
    \caption{Comparision results (NDCG@10) on retrieval tasks.}
    \resizebox{0.95\hsize}{!}{
    \begin{tabular}{lcc}
        \toprule
        \textbf{Methods/Datasets} & \textbf{SciFact} & \textbf{FIQA} \\
        \midrule
        \multicolumn{1}{l}{\textit{\textbf{Language Models Only}}} \\
        BERT-Base  \cite{DBLP:conf/naacl/DevlinCLT19} & 13.3 &   2.2 \\
        RoBERTa-Large  \cite{DBLP:journals/corr/abs-1907-11692} & 43.3 & 20.4 \\
        E5-Small \cite{wang2022text} & 65.6 &  34.8 \\
        E5-Base \cite{wang2022text} & 73.1 & 36.4 \\
        E5-Large \cite{wang2022text} & 72.6 &  38.6 \\
        GTR-XXL \cite{ni2022large} & 66.2 & 46.7\\
        SGPT \cite{muennighoff2022sgpt} & 74.7 & 37.2 \\
        E5-Mistral, 0-shot \cite{wang2023improving} & 76.1 & 53.5 \\
        E5-Mistral, LoRA \cite{wang2023improving} & 75.9 & 53.2  \\
        \midrule
        \multicolumn{2}{l}{\textit{\textbf{LM + KG}}} \\
        QA-GNN \cite{DBLP:conf/naacl/YasunagaRBLL21} & 41.4 & 19.5 \\
        GreaseLM \cite{zhang2022greaselm} & 48.9 & 29.3 \\
        \midrule
        {\ourmethod, LP (Ours)} & \textbf{82.9} & \underline{54.1}  \\
        {\ourmethod, LoRA (Ours)} & \underline{80.8}  & \textbf{56.2}  \\
        \bottomrule
    \end{tabular}
    }
    \vspace{-1em}
    \label{tab:retrieval}
\end{table}

% $\pm$ 1.9
% $\pm$ 0.1
% $\pm$ 0.9
% $\pm$ 0.3
% When comparing \ourmethod with E5-Mistral, the backbone of our model and the best-performing baseline, we observe a clear improvement on both datasets. This performance boost highlights the effectiveness of incorporating graph-based contextual information into the retrieval process.

\begin{figure*}[t]
    \centering
    \includegraphics[width=0.9\linewidth]{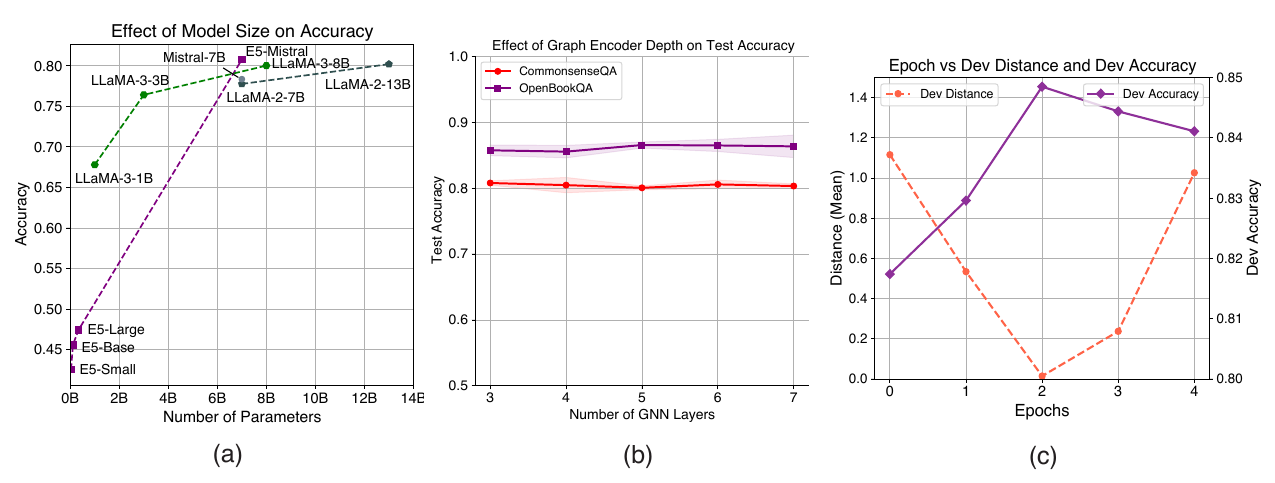}
    \vspace{-1em}
    \caption{(a) The effect of LLM backbone choice on accuracy for the CommonsenseQA dataset. The figure shows three series: E5, LLaMA-3, and LLaMA-2, along with a single Mistral-7B model. (b) The effect of graph encoder depth (number of GNN layers) on test accuracy for CommonsenseQA and OpenBookQA datasets. The shaded areas represent the standard deviation, indicating the variance in performance across different trials. (c) Graph-text embedding distance (red dashed) and dev accuracy (purple solid) on CommonsenseQA across training epochs.}
    \label{fig:combine}
    \vspace{-1em}
\end{figure*}

As shown in Table \ref{tab:retrieval}, \ourmethod achieves the best performance on both datasets. A closer examination of the LM + KG baselines reveals interesting trends. Both models use RoBERTa-Large in their architectures. From the results, GreaseLM outperforms the RoBERTa-Large model by a margin of 5.6\% on SciFact and 8.9\% on FiQA, which aligns with our earlier findings that graph context can be useful for enhancing retrieval performance. However, QA-GNN shows a performance degradation compared to RoBERTa-Large, particularly on FiQA (19.5 vs. 20.4). The reason for this drop may be attributed to QA-GNN’s use of a simple MLP for combining the graph and text embeddings. This weaker integration mechanism is likely insufficient for fully leveraging the graph context, resulting in suboptimal performance. In contrast, GreaseLM employs a shallow transformer, which offers better capacity for fusing multimodal information, leading to moderate gains.

% Our method, \ourmethod, demonstrates its superiority by using a powerful LLM backbone to combine the graph and text representations, which is reflected in the substantial performance improvements on both datasets. This highlights the advantage of using a more expressive model to combine graph and textual information.

\subsection{Ablation Studies}
\label{subsec:ablation}
To better understand the contributions of different components in \ourmethod, we perform an ablation study on the CommonsenseQA and OpenBookQA datasets under the linear probe setting. More results can be found in Appendix \ref{sec:appendix_results}.

% The results, presented in Table \ref{tab:ablation}, highlight the impact of key components on overall performance. Additionally, Figure \ref{fig:combine} and Table \ref{tab:emb_init} show the effects of different choices of components on model performance.

% combine the ablation study tables
% how to choose baselines
% 

% \begin{figure}
%     \centering
%     \includegraphics[width=0.85\linewidth]{figures/model_size.pdf}
%     \caption{The effect of LLM backbone choice on accuracy for the CommonsenseQA dataset. The figure shows three series: E5, LLaMA-3, and LLaMA-2, along with a single Mistral-7B model.}
%     \label{fig:llm_backbone}
% \end{figure}

\noindent\textbf{Effect of LLM Backbone Choice.} \ourmethod exhibits flexibility in adopting various LLMs as its backbone. To evaluate the impact of different LLM backbones on performance, we compare four series of models: the E5 series \cite{wang2022text, wang2023improving}, LLaMA-2 \cite{touvron2023llama}, LLaMA-3 \cite{dubey2024llama}, and Mistral \cite{jiang2023mistral}, as illustrated in Figure \ref{fig:combine}(a). We first observe that increasing the model size within each series consistently enhances the performance. This trend highlights that larger models, with more parameters, are typically able to better capture complex relationships in multi-modal data, leading to higher performance. Additionally, LLaMA-3 significantly improves over LLaMA-2. Specifically, the performance of LLaMA-3-3B matches that of LLaMA-2-7B and LLaMA-3-8B achieves results comparable to LLaMA-2-13B, indicating that the new LLaMA-3 architecture brings considerable advancements compared to its predecessor LLaMA-2.

% Furthermore, when comparing E5-Mistral to the base Mistral model, we see that E5-Mistral, which is a fine-tuned version of Mistral optimized for text embedding tasks, outperforms Mistral by a noticeable margin. This indicates that models specifically tailored for text embeddings are better suited for tasks involving joint embeddings of text and graph data.

% \begin{figure}
%     \centering
%     \includegraphics[width=0.75\linewidth]{figures/gnn_layer.pdf}
%     \caption{The effect of graph encoder depth (number of GNN layers) on test accuracy for CommonsenseQA and OpenBookQA datasets. The shaded areas represent the standard deviation, indicating the variance in performance across different trials.}
%     \label{fig:graph_layer_num}
% \end{figure}

% The darker lines in the figure represent the mean test accuracy, while the shaded regions around the lines indicate the standard deviation. 

\noindent \textbf{Effect of Graph Encoder Depth.} We evaluate the effect of varying the number of GAT layers in the graph encoder on the performance of \ourmethod using the CommonsenseQA and OpenBookQA datasets. As shown in Figure \ref{fig:combine}(b), both datasets exhibit slight fluctuations in test accuracy as the number of GAT layers increases from 3 to 7. This suggests that the method is quite robust to changes in GAT depth, with only small variations in performance.

% For CommonsenseQA, the accuracy remains relatively stable across different GNN depths, fluctuating around 80\%, with minor drops at 5 and 7 layers. This suggests that the method is quite robust to changes in GAT depth, with only small variations in performance. Similarly, for OpenBookQA, we also observe a stable trend, with accuracy consistently around 86\%.

\noindent \textbf{Effect of Graph-Text Alignment.} We further investigate the effectiveness of graph-text alignment by studying how the alignment evolves during training. Specifically, we calculate the mean Euclidean distance between the normalized graph and text embeddings on the dev set (shown as the red dashed line) and compare it to the corresponding development accuracy (purple solid line) across different training epochs, as illustrated in Figure \ref{fig:combine}(c). This figure shows a clear inverse relationship between the two curves: as the graph-text distance decreases, accuracy improves. Notably, at epoch 2, the graph-text distance reaches its minimum, while the dev accuracy peaks. This trend suggests that better alignment between graph and text embeddings contributes directly to improved model performance, highlighting the effectiveness of our graph-text alignment strategy in \ourmethod.
\section{Conclusion}
In this paper, we introduce \ourmethod, a simple yet effective framework designed to integrate graph and text data using a novel alignment strategy and LLMs. We have demonstrated that \ourmethod enhances the semantic coherence between these two modalities, resulting in significantly improved performance on several NLP tasks and datasets. By aligning graph embeddings directly with text embeddings, \ourmethod ensures a deeper integration of structured and unstructured data. 

 % (QA, graph-text pair classification, and retrieval)

\section*{Limitations}
While our proposed \ourmethod achieves strong performance across multiple tasks, it has certain limitations. First, our approach does not explicitly handle additional modalities such as images, which could provide richer contextual signals. Exploring effective ways to incorporate multi-modal data remains an important direction for future research. Second, although our model effectively aligns graph and text representations, there may be further opportunities to refine this alignment for better generalization across diverse tasks and domains.

% Future directions include exploring additional ways to fine-tune the alignment mechanisms to further enhance model performance, and investigate the scalability of our approach in handling larger datasets.

\bibliography{main}

\newpage
\appendix
\section{Ethical Use of Data and Informed Consent}
All datasets used in this work are publicly available and widely used in the research community. No personally identifiable information (PII) or sensitive data is included in these datasets. Our research strictly adheres to the ethical guidelines of using publicly available datasets, and no additional data was collected from human subjects. 

\section{Related Work}
\noindent \textbf{Graphs and Language Models for Multi-Modal Embedding Tasks.}
Integration of graph data with language models has been an evolving field, aiming to combine structured graph data with unstructured textual information for enhanced data analysis. Historically, early attempts in this area employed an MLP or a shallow transformer to merge the information from both modalities \cite{lin2019kagnet, DBLP:conf/naacl/YasunagaRBLL21, zhang2022greaselm, DBLP:conf/emnlp/FengCLWYR20}, which may not fully exploit the rich contextual information from text and graph data. 

The advent of LLMs has brought a transformative shift to the NLP field.  LLMs, with their extensive pre-training on diverse corpora, offer unprecedented capabilities for deep semantic understanding and reasoining \cite{zhao2023survey, gao2023examining}. Recent research has begun to explore the potential of LLMs for enhancing multimodal embedding tasks of graphs and text \cite{huang2024gnns,tian2024graph, zhao2023graphtext, he2024harnessing, zhu2024parameter}. These studies primarily focus on augmenting the graph representation capabilities within the LLM framework. While these efforts mark significant advancements, they predominantly concentrate on graph representation learning rather than direct NLP tasks. Additionally, these methods do not explicitly align the semantic spaces of graphs and text. In contrast, our approach not only integrates LLMs for handling complex multimodal inputs but also introduces an explicit alignment mechanism between graph and text embeddings. This alignment is crucial for enhancing the semantic integration and boosting performance across diverse NLP tasks by capturing complex intermodal relationships.

There are recent studies that have explored using LLMs and graph data for generative tasks beyond embedding tasks \cite{he2024g, perozzi2024let, zhang2024graphtranslator, chai2023graphllm, hu2024grag, DBLP:conf/acl/JinXZRZL0TWM024, zhu2024investigating}. Our work, however, concentrates on embedding tasks, which is an orthogonal direction. Unlike generative tasks, which primarily focus on creating new content via next-token prediction, embedding tasks are aimed at developing rich, informative representations that can be used directly to enhance performance in downstream applications such as classification and retrieval.
\vspace{0.5em}

% Recent works have explored the integration of graphs with language models to leverage both structured graph data and unstructured textual information \cite{lin2019kagnet, DBLP:conf/naacl/YasunagaRBLL21, zhang2022greaselm, DBLP:conf/emnlp/FengCLWYR20}. These studies typically employ a MLP or a shallow transformer to merge the information from both modalities. However, these approaches often face limitations in capacity of the modules for modality integration, which may not fully exploit the rich contextual information from text and graph data. 

% In contrast, our approach utilizes LLMs, which offer a deeper understanding of context. By employing LLMs, we can more effectively process and interpret the rich, interconnected data presented by graphs and text. This allows for a more robust and nuanced handling of multimodal inputs, leading to superior performance on embedding tasks.

% In our method, we extend this line of work by introducing a more sophisticated approach to integrating graph embeddings with large language models (LLMs). While prior models often rely solely on GNN backbones for graph processing, our approach adopts a more powerful fusion technique where graph embeddings are aligned with text embeddings through the use of a large-scale language model, resulting in improved performance on tasks that require both graph and text understanding.
\noindent \textbf{Contrastive Learning.}
Contrastive learning has gained significant attention in recent years as a method to learn representations by maximizing agreement between positive pairs while pushing negative pairs apart in the embedding space \cite{he2020momentum, chen2020simple, linpisces, DBLP:conf/iclr/LinXXW24, yu2022graph}. Pioneering works like SimCLR \cite{chen2020simple} and MoCo \cite{he2020momentum} have demonstrated the effectiveness of contrastive learning in the visual domain. Similar approaches have been applied to graphs \cite{yu2022graph, you2020graph, fang2024gaugllm} and language models \cite{iter2020pretraining, fang2020cert, giorgi2021declutr}, but these methods typically operate within a single modality (e.g., graphs or text). In contrast, our approach introduces contrastive learning to align the graph embedding space with the text embedding space. By transforming graphs into textual descriptions and applying contrastive learning, we ensure that embeddings from both modalities align closely, enhancing the model's ability to perform tasks that require both graph-based reasoning and text comprehension.

\section{Dataset Details}
\subsection{KG-Contextulized QA}
\label{subsec:app_kqqa}
\textbf{CommonsenseQA} \cite{DBLP:conf/naacl/TalmorHLB19} is a 5-way multiple-choice QA dataset focused on applying commonsense knowledge in answering questions. It includes 12,102 questions, each with one correct answer and four distractor answers. 

\textbf{OpenBookQA} \cite{DBLP:conf/emnlp/MihaylovCKS18} is a 4-way multiple choice QA dataset that requires reasoning with elementary science knowledge, containing 5,957 questions.

\textbf{MedQA-USMLE} \cite{DBLP:journals/corr/abs-2009-13081} is a 4-way multiple choice QA task that requires biomedical and clinical knowledge. The questions are originally from practice tests for the United States Medical License Exams (USMLE). The dataset contains 12,723 questions.

For each question in the QA datasets, a subgraph context extracted from a KG is utilized to provide additional contextual information, following Yasunaga et al. \cite{DBLP:conf/naacl/YasunagaRBLL21}. Specifically, for CommonsenseQA and OpeBookQA, we used the ConceptNet \cite{DBLP:conf/aaai/SpeerCH17}, which contains 799,273 nodes and 2,487,810 edges. Node embeddings are initialized by Roberta-Large \cite{DBLP:journals/corr/abs-1907-11692} and kept frozen during the training process, following \citet{DBLP:conf/naacl/YasunagaRBLL21}. The query node embeddings mentioned in Section \ref{subsec:graph_data_encoding} are calculated by the LLM backbone. For MedQA-USMLE dataset, we used the UMLS knowledge graph used in \citet{DBLP:conf/naacl/YasunagaRBLL21}, which contains 9,958 nodes and 44,561 edges. Node embeddings are initialized using SapBERT \cite{DBLP:conf/naacl/LiuSMBC21}, following \citet{DBLP:conf/naacl/YasunagaRBLL21}. 

\subsection{Graph-Text Pair Classification}
\label{appendix:subsec_webnlg}
To evaluate \ourmethod on graph-text pair classification, we curated a dataset based on \textbf{WebNLG} \cite{web_nlg}. The original WebNLG dataset is used to assess models' ability in text-to-graph and graph-to-text generation. Concretely, each data contains a graph-text pair where the graph is a set of triples from DBpedia and the text is the corresponding description of the triples. For example, the graph data are \textit{"(John\_E\_Blaha birthDate 1942\_08\_26), (John\_E\_Blaha birthPlace San\_Antonio), (John\_E\_Blaha occupation Fighter\_pilot)"}, while the corresponding text is \textit{"John E Blaha, born in San Antonio on 1942-08-26, worked as a fighter pilot."}

In this paper, we use the v3.0 release data to construct the dataset for graph-text pair classification. First, we curated the dataset by identifying relations that appear in the training, dev, and test sets. We then filtered the dataset to retain only those graph triples that contain these relations for all three sets. To further increase the complexity of the task, we generated a series of new, randomly combined positive samples. Specifically, we merged multiple graph-text pairs from the original dataset, creating data with larger graphs and longer text by concatenating their respective triples and sentences. 

Next, we generated an equal number of negative samples. To this end, we followed a similar process to create positive pairs but introduced mismatches. In this case, while the graph triples were taken from one pair, the text was taken from a different, unrelated pair. This ensured that the graph and text did not align, resulting in non-matching pairs that serve as negative examples. We finally constructed 10,000 training, 4,000 validation, and 2,000 test data.
\subsection{Retrieval}
% use 100 of 890 for validation set

\textbf{SciFact} \cite{DBLP:conf/emnlp/WaddenLLWZCH20} is a scientific fact-checking dataset aimed at verifying scientific claims using relevant research papers. It contains 920 training queries, and 300 test queries, with a corpus of 5,183 documents. In our experiment, we split 100 samples from the training set to serve as validation queries, ensuring that the remaining training data is used for model training while still providing a separate validation set for tuning and evaluation.

\textbf{FiQA} \cite{DBLP:conf/www/MaiaHFDMZB18} is a financial-domain retrieval dataset designed to address complex financial question answering and information retrieval tasks. It contains 14,166 training queries, 500 development queries, and 648 test queries, with a total of 57,638 documents in the corpus.

\section{Methodology Details}
\subsection{Additional Framework Details}
% Dual-View
\ourmethod utilizes a dual-view architecture for graph-text alignment, incorporating two parallel branches. The first branch processes the original graph together with the text, while the second branch processes a textual description of the graph along with the input text. In our experiments, both branches are employed to generate embeddings that are used for downstream tasks, and both contribute to the task-specific loss $\mathcal{L}^{\rm (task)}$. This dual-branch approach has several advantages: it allows the model to learn complementary information from both the structured graph and its textual description, enhancing the overall representation. During the evaluation, however, we simplify the process by using only the graph+text branch, and it has shown to provide sufficient performance for the downstream tasks, eliminating the need for the graph description branch.

\subsection{Training Objective}
\label{appendix:training_objective}
In this part, we detail the task-specific loss function $\mathcal{L}^{\rm (task)}$ for each downstream task.

\subsubsection{KG-Contextualized QA}
For the KG-contextualized QA task, we apply the cross-entropy loss function, which is crucial for classification tasks involving multiple-choice questions. This loss function is computed as follows:
\begin{align}
    \mathcal{L}^{\rm (task)} = - \sum_{i=1}^{n_{\rm batch}} \sum_{j=1}^{n_{\rm choice}} y_j^{(i)} \log(\hat{y}_j^{(i)})
\end{align}
where $n_{\rm batch}$ is the number of samples in a batch and $n_{\rm choice}$ is the number of answer choices per question. $y_j^{(i)}$ is a binary indicator (1 if the choice $j$ is the correct answer for sample $i$, and 0 otherwise). $\hat{y}_j^{(i)}$ is the predicted probability that choice $j$ is the correct answer for sample $i$. 

% We first evaluated our Janus framework on KG-contextualized QA tasks. Each data point for KG-contextualized QA tasks contains a question and candidates. We train our \ourmethod framework using a cross-entropy loss function. 

% We retrieve a relevant sub-graph for each candidate, following \cite{DBLP:conf/naacl/YasunagaRBLL21, zhang2022greaselm}. Note that, sub-graph retrieval is not the main focus of this paper and we just follow the previous work and put more attention on the subsequent parts. We use the graph encoder mentioned previously to obtain graph embeddings of each sub-graph. 

\subsubsection{Graph-Text Pair Classification Tasks}
In the graph-text pair classification task, we use binary cross-entropy loss to determine the match/mismatch status between the graph and text pairs:
\begin{align}
\resizebox{0.96\hsize}{!}{$
    \mathcal{L}^{\rm (task)} = - \sum_{i=1}^{n_{\rm batch}} \left(y^{(i)}\log \hat{y}^{(i)} + (1 - y^{(i)} \log (1-\hat{y}^{(i)})\right)
    $}
\end{align}
Here, $y^{(i)}$ is the true label (1 if the pair matches, 0 otherwise) and $\hat{y}^{(i)}$ is the predicted probability of a match as output by the MLP classifier.

\subsubsection{Retrieval Tasks}
For retrieval tasks, we apply infoNCE loss \cite{oord2018representation} for training. Specifically, given a batch of positive pairs $(d_i, p_i)$, we assume that $(d_i, p_i)$ is a positive pair and $(d_i, p_j)$ for $i \neq j$ a negative pair. By applying infoNCE loss, we have
\begin{align}
\resizebox{0.96\hsize}{!}{$
    \mathcal{L}^{(task)} = -\sum_{i=1}^{n_{\text{batch}}} \log \left( \frac{e^{\text{sim}(a_i, p_i)/\tau}}{e^{\text{sim}(a_i, p_i)/\tau} + \sum_{j \neq i} e^{\text{sim}(a_i, p_j)/\tau}} \right)$
    }
\end{align}
where $\text{sim}(\cdot, \cdot)$ is the consine similarity between the embeddings. $\tau$ is the temperature scaling parameter.

\begin{table*}
\centering
\caption{Ablation Study on different branches of \ourmethod: This table reports the test accuracy of \ourmethod when using only the graph + text branch or only the graph description + text branch.}
\resizebox{0.6\hsize}{!}{
\begin{tabular}{lcc}
\toprule
\textbf{Models}         & \textbf{CommonsenseQA} & \textbf{OpenBookQA}   \\ 
\midrule
GreaseLM \cite{zhang2022greaselm} & 74.20 $\pm$ 0.40 & 83.87 $\pm$ 1.29 \\
\midrule
Only graph + text & 79.69 $\pm$ 0.85 & 84.80 $\pm$ 1.06 \\
Only description + text & 78.32 $\pm$ 0.61  & 84.00 $\pm$ 0.92    \\
{\ourmethod} & \textbf{81.09} $\pm$ 0.73 & \textbf{86.67} $\pm$ 1.10 \\
\bottomrule
\end{tabular}
}
\label{tab:eff_multi_branch}
\end{table*}

\begin{table*}[t]
\centering
\caption{The effect of graph encoder choice on test accuracy.}
\resizebox{0.6\hsize}{!}{
\begin{tabular}{lcc}
\toprule
\textbf{Graph Encoder}         & \textbf{CommonsenseQA}  & \textbf{OpenBookQA}  \\ 
\midrule
GCN  \cite{DBLP:conf/iclr/KipfW17} & 80.95 $\pm$ 0.28  &  84.67 $\pm$ 1.27  \\
GAT \cite{DBLP:conf/iclr/VelickovicCCRLB18}      & 81.09 $\pm$ 0.73 &  86.67 $\pm$ 1.10 \\
GIN \cite{DBLP:conf/iclr/XuHLJ19}       & 80.23 $\pm$ 0.57  & 85.40 $\pm$ 0.92   \\
\bottomrule
\end{tabular}
}
\label{tab:gnn_encoder}
\end{table*}

\begin{table}[h]
    \centering
    \caption{LoRA Hyperparameters}
    \label{tab:lora_params}
    \resizebox{0.9\hsize}{!}{
    \begin{tabular}{lc}
        \toprule
        \textbf{Hyperparameter} & \textbf{Value} \\
        \midrule
        Rank ($r$) & 8 \\
        Alpha ($\alpha$) & 16 \\
        Dropout & 0.05 \\
        Target Modules & \{q\_proj, k\_proj, v\_proj, o\_proj, \\
        & gate\_proj, up\_proj, down\_proj\} \\
        \bottomrule
    \end{tabular}
    }
\end{table}

\section{Experiment Details}
\label{app:section:exp}
\begin{table*}
\centering
\caption{Hyperparameters for training \ourmethod.}
\resizebox{\hsize}{!}{
\begin{tabular}{lcccccc}
\toprule
\textbf{Hyperparameter}         & \textbf{CommonsenseQA}  & \textbf{OpenBookQA} & \textbf{MedQA-USMLE} & \textbf{WebNLG} & \textbf{SciFact} & \textbf{FiQA}  \\ 
\midrule
GNN hidden dim & 256 & 256 & 256 & 256 & 256 & 512 \\
Number of GNN layers & 3 & 5 & 5 & 5 & 3 & 3\\
GAT attention heads & 2 & 2 & 2 & 2 & 2 &2 \\
Dropout rate & 0.2 & 0.2 & 0.2 & 0.2 & 0.2 & 0.2\\
Context length & 128 & 128 & 256 & 256 & 512 & 512 \\
\midrule
Learning rate & $1\times 10 ^{-5}$ & $1\times 10 ^{-5}$ & $1\times 10 ^{-5}$ & $1\times 10 ^{-4}$ & $1\times 10 ^{-5}$ & $5\times 10 ^{-6}$\\
Optimizer & RAdam & RAdam & RAdam & RAdam & RAdam & RAdam\\
Weight decay & $1\times 10^{-2}$ & $1\times 10^{-2}$ & $1\times 10^{-2}$ & $1\times 10^{-2}$ & $1\times 10^{-2}$ & $1\times 10^{-2}$\\
Learning rate schedule & constant & constant  & constant  & constant  & constant  & constant\\
Number of epochs & 5 & 5 & 15 & 5 & 15 & 1 \\
Batch size & 8 & 8 & 8 & 32 & 16 & 256 \\
Max gradient norm & 1.0 & 1.0 & 1.0 & 1.0 & 1.0 & 1.0 \\
$\lambda$ in Eq. (\ref{eq:loss_comb}) & 0.05 & 0.05 & 0.05 & 0.05 & 0.05 & 0.05\\
\midrule
Knowledge graphs & ConceptNet & ConceptNet & UMLS & DBPedia & ConceptNet & ConceptNet \\
Max number of nodes in subgraphs & 200 & 200 & 200 & 200 & 200 & 200\\
Number of relations & 38 & 38 & 34 & 748 & 38 & 38 \\
\bottomrule
\end{tabular}
}
\label{tab:hyperparams}
\end{table*}

\subsection{Additional Implementation Details}
We implement models using PyTorch \cite{paszke2019pytorch} and PyTorch Geometric \cite{fey2019fast}. All experiments are conducted on a single A100 80GB GPU. For baselines that we implemented ourselves, we followed the settings and hyperparameters described in the original papers to ensure fair comparisons. Detailed hyperparameter settings for \ourmethod can be found in Table \ref{tab:lora_params} and Table \ref{tab:hyperparams}.

% Additionally, since we freeze the parameters of the LLM backbone in \ourmethod, the graph descriptions are encoded in advance before training. This strategy improves training efficiency and significantly reduces memory usage. For the E5-Mistral \cite{wang2023improving} baseline, we freeze the parameters and only train the classification head, while for other LMs we update all parameters. 

\subsection{Additional Results}
\label{sec:appendix_results}
\subsubsection{Effect of graph context information.}When \ourmethod is input with only textual input, excluding graph context information, there is a significant drop in performance, with accuracy decreasing to 69.49\% on CommonsenseQA and 74.80\% on OpenBookQA. This sharp decline underscores the critical role that graph context plays in enhancing the model’s ability to understand and reason with the data. The graph provides structured knowledge that complements the unstructured text, and without it, the model relies solely on the textual input, which limits its capacity to effectively handle complex embedding tasks.

\subsubsection{Effect of Graph-Text Alignment.}  Removing contrastive learning from \ourmethod leads to a performance drop, with the accuracy decreasing from 81.09\% to 79.69\% on CommonsenseQA, and from 86.67\% to 84.80\% on OpenBookQA. This highlights the importance of contrastive learning in aligning graph and text embeddings, ensuring better integration of multimodal information.

\begin{table*}[t]
    \centering
    \caption{Ablation study results on test accuracy.}
    \resizebox{0.6\hsize}{!}{
    \begin{tabular}{lcc}
        \toprule
        \textbf{Methods} & \textbf{CommonsenseQA} & \textbf{OpenBookQA} \\
        \midrule
        GreaseLM \cite{zhang2022greaselm} & 74.20 $\pm$ 0.40 & 83.87 $\pm$ 1.29 \\
        \midrule
        {\ourmethod} w/o graph & 69.49 $\pm$ 0.28 & 74.80 $\pm$ 0.35 \\
        {\ourmethod} w/o alignment & 79.69 $\pm$ 0.85 & 84.80 $\pm$ 1.06 \\
        {\ourmethod} & \textbf{81.09} $\pm$ 0.73 & \textbf{86.67} $\pm$ 1.10 \\
        \bottomrule
    \end{tabular}
    }
    \label{tab:ablation}
    \vspace{-1em}
\end{table*}

\subsubsection{Effect of Different Branches in \ourmethod}
To better understand the contributions of each branch in \ourmethod, we perform an ablation study, as shown in Table \ref{tab:eff_multi_branch}. This analysis evaluates the performance of \ourmethod when using only the graph + text branch or only the graph description + text branch and compares it to the full version of \ourmethod, which integrates both branches. We find that while each branch individually yields comparable performance, the combination of both branches—enhanced by our graph-text alignment technique—leads to significant performance gains. This demonstrates the effectiveness of our proposed methods. 

\subsubsection{Effect of Graph Encoder Choice}
We also investigate the impact of different graph encoders on the performance of \ourmethod, as shown in Table \ref{tab:gnn_encoder}. We observe that the performance with GAT achieves the best results on both CommonsenseQA and OpenBookQA.

\subsubsection{Impact of Node Embedding Initialization}

\begin{table*}[h]
\centering
\caption{Impact of different graph node embedding initialization models on accuracy for the CommonsenseQA dataset. }
\resizebox{0.6\hsize}{!}{
\begin{tabular}{lcc}
\toprule
\textbf{Model}         & \textbf{Output Dimension}  & \textbf{Accuracy}  \\ 
\midrule
RoBERTa-Large  \cite{DBLP:journals/corr/abs-1907-11692} & 1,024 & 81.09 $\pm$ 0.73   \\
E5-Small \cite{wang2022text}      & 384 & 80.37 $\pm$ 0.38  \\
E5-Base \cite{wang2022text}       & 768 & 80.77 $\pm$ 0.05   \\
E5-Large \cite{wang2022text}      & 1,024  & 80.79 $\pm$ 0.30   \\ 
E5-Mistral \cite{wang2023improving}    & 4,096 & 80.42 $\pm$ 0.90  \\
\bottomrule
\end{tabular}
}
\label{tab:emb_init}
\end{table*}

To assess the impact of different node embedding initialization models on performance, we conducted experiments on the CommonsenseQA dataset. As shown in Table \ref{tab:emb_init}, the results reveal minimal differences in accuracy across models with varying output dimensions, indicating that the choice of initialization model has limited influence on performance. For instance, RoBERTa-Large, E5-Base, and E5-Large achieve nearly identical results, with an accuracy of around 81\%.

Interestingly, even when using lighter models for node embedding initialization, such as E5-Small, which has a smaller output dimension (384), the performance remains competitive. This demonstrates that even when the node initialization model differs from the actual backbone used in \ourmethod, there is no significant performance degradation. For example, compared with E5-Mistral node embedding initialization, the lighter E5-Small model offers the advantage of reduced computational cost without compromising much on accuracy. 

% \subsubsection{KG-Contextulized QA}
% We adopt E5-Mistral \cite{wang2023improving} as \ourmethod's LLM backbone. The number of GNN layers is 3 for CommonsenseQA and 5 for the other two datasets. The hidden size of GNNs is set to 256. We train \ourmethod using RAdam optimizer \cite{DBLP:conf/iclr/LiuJHCLG020} with a weight decay of $1\times 10^{-2}$. The batch size is 8 for CommonsenseQA and OpenBookQA with 5 epochs, and 8 for MedQA-USMLE with 15 epochs. The learning rate is set $1\times 10^{-5}$. The max lengths of input tokens are set 256 for MedQA-USMLE and 128 for the other two datasets. Additionally, we selected the weight values $\lambda = 0.05$ for the loss function combination.

% \subsubsection{Graph-Text Pair Classification}
% In our experiments, the LLM backbone of \ourmethod is E5-Mistral. The GNN layer num is 5 and hidden dimension is 256. The learning rate is $1\times 10^{-4}$ with a batch size of 32 and 5 epoch training. Additionally, we set the weight values as $\lambda = 0.05$.

% \subsubsection{Retrieval}
% % QAGNN + Greaslm modified
% Additionally, we selected the weight values $\lambda = 0.05$ for the loss function.

\end{document}